\documentclass[10pt,twocolumn,letterpaper]{article}

\usepackage[pagenumbers]{stylefile}

\usepackage[dvipsnames]{xcolor}

\usepackage[english]{babel}
\usepackage{graphicx}
\usepackage{amsmath}
\usepackage{amssymb}
\usepackage{booktabs}
\usepackage{multicol}
\usepackage{multirow}
\usepackage{stfloats}
\usepackage{tikz}
\usepackage{algorithm}
\usepackage{algorithmicx}
\usepackage{algpseudocode}
\usepackage{colortbl}
\usepackage{listings}
\definecolor{codegreen}{rgb}{0,0.6,0}
\definecolor{codegray}{rgb}{0.5,0.5,0.5}
\definecolor{codepurple}{rgb}{0.58,0,0.82}
\definecolor{backcolour}{rgb}{1,1,1}
\definecolor{orangebackcolor}{rgb}{1,0.9,0.8}

\definecolor{teaser-green}{RGB}{0, 102, 0}
\definecolor{teaser-orange}{RGB}{229, 134, 0}
\definecolor{teaser-grey}{RGB}{111, 111, 111}
\definecolor{teaser-blue}{RGB}{52, 47, 215}

\definecolor{myblue}{rgb}{0.21,0.49,0.74}
\usepackage[pagebackref,breaklinks,colorlinks,citecolor=myblue]{hyperref}

\definecolor{tabfirst}{rgb}{1, 0.7, 0.7} %
\definecolor{tabsecond}{rgb}{1, 0.85, 0.7} %
\definecolor{tabthird}{rgb}{1, 1, 0.7} %

\usepackage[capitalize]{cleveref}
\crefname{section}{Sec.}{Secs.}
\Crefname{section}{Section}{Sections}
\Crefname{table}{Table}{Tables}
\crefname{table}{Tab.}{Tabs.}
\Crefname{figure}{Figure}{Figures}
\crefname{figure}{Fig.}{Figs.}

\begin{document}

\title{Unsupervised Anomaly Detection using Aggregated Normative Diffusion}
\author{
    Alexander Frotscher\thanks{Equal contribution.}~~~~~~Jaivardhan Kapoor$^{*}$~~~~~~Thomas Wolfers~~~~~~Christian F. Baumgartner\\
    University of Tübingen\\
}

\maketitle

\begin{abstract}

Early detection of anomalies in medical images such as brain MRI is highly relevant for diagnosis and treatment of many conditions. Supervised machine learning methods are limited to a small number of pathologies where there is good availability of labeled data. In contrast, \emph{unsupervised} anomaly detection (UAD) has the potential to identify a broader spectrum of anomalies by spotting deviations from normal patterns.
Our research demonstrates that existing state-of-the-art UAD approaches do not generalise well to diverse types of anomalies in realistic multi-modal MR data. 
To overcome this, we introduce a new UAD method named Aggregated Normative Diffusion (\texttt{ANDi})\footnote{Code is available at \href{https://github.com/alexanderfrotscher/ANDi}{https://github.com/alexanderfrotscher/ANDi}.}. \texttt{ANDi} operates by aggregating differences between predicted denoising steps and ground truth backwards transitions in Denoising Diffusion Probabilistic Models (DDPMs) that have been trained on pyramidal Gaussian noise. 
We validate \texttt{ANDi} against three recent UAD baselines, and across three diverse brain MRI datasets. We show that \texttt{ANDi}, in some cases, substantially surpasses these baselines and shows increased robustness to varying types of anomalies.  Particularly in detecting multiple sclerosis (MS) lesions, \texttt{ANDi} achieves improvements of up to 178\% in terms of AUPRC.
\end{abstract}
\section{Introduction}
\label{sec:intro}
Detection of anomalies in medical imaging data is highly important for early diagnoses and timely treatment of diseases. In order to detect anomalies, clinicians typically evaluate \emph{multi-modal} image data, consisting of a range of different medical images of the same patient that provide different information about potential pathologies. In neuroimaging, routinely, a range of different magnetic resonance (MR) imaging sequences are acquired that together offer a comprehensive view of the brain. 

\begin{figure}[ht!]
    \centering
    \includegraphics[width=0.99\linewidth]{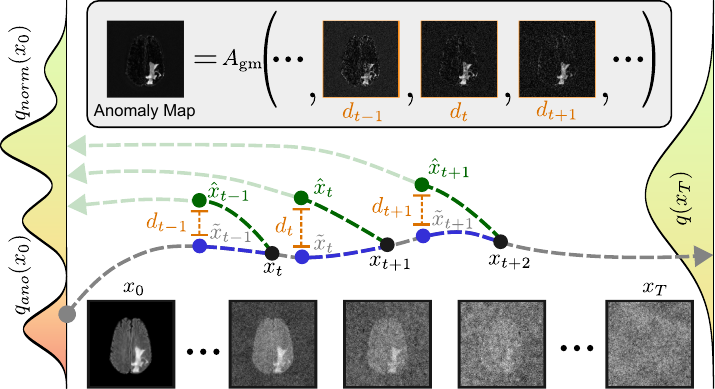}
    \caption{\textbf{Aggregated Normative Diffusion (\texttt{ANDi})}. First, we train a DDPM model using our proposed Gaussian pyramidal noise on healthy brain slices to approximate the normative distribution $q_{norm}(x_0)$. In order to obtain an anomaly map for a possibly anomalous image $x_0$, we first partially noise it using the Gaussian forward process (indicated by the gray arrow). We then calculate the pixel-wise \textcolor{teaser-orange}{Euclidean distance $d_t$} between the \textcolor{teaser-blue}{ground truth backwards transition $\tilde{x}_t = \mu_q(x_t, x_0)$} and the \textcolor{teaser-green}{denoising step $\hat{x}_t = \mu_\theta(x_t, t)$} for a partial range of $t$. The denoising step towards \textcolor{teaser-green}{$\hat{x}_t$} can be thought of as \emph{normative diffusion} as it is taking one step towards the normative distribution. Finally, using the geometric mean $A_{\text{gm}}$, we aggregate deviations over \textcolor{teaser-orange}{$d_t$}.}
    \label{fig:teaser}
\end{figure}

Due to the diversity of possible anomalies, their detection is very challenging, time-consuming, and prone to errors. Therefore, there has been substantial interest in automating anomaly detection. The majority of research effort on automated anomaly detection has been focused on supervised learning (\eg \cite{brats21,hernandez2022isles,shifts2021,liew2022large}). However, supervised algorithms must be tailored to each type of anomaly and hinge on large-scale hand-annotated data. This limits their clinical applicability and prevents their application to rare or understudied diseases. 

Recently, there has been an increased interest in \textit{unsupervised} anomaly detection (UAD) methods. UAD attempts to learn a concept of normality and label anything that deviates from normality as an anomaly. Recent methods often apply generative modeling to learn a \emph{normative reference distribution} from a large-scale population of healthy images. To this end, prior work has employed techniques such generative adversarial networks (GANs)~\cite{f-anogan,schlegl2017unsupervised}, autoencoders (AEs)~\cite{baur2020bayesian,sato2018primitive}, variational autoencoders (VAEs)~\cite{bercea2023generalizing,bercea2023reversing,normative-ascent-elbo}, and most recently denoising diffusion probabilistic models (DDPMs)~\cite{anoddpm, autoddpm, pinaya2022fast}. Most prior work then uses the trained networks to reconstruct a pseudo-healthy image from an input image potentially containing an anomaly~\cite{bercea2023reversing,normative-ascent-elbo,dae,autoddpm,pinaya2022fast}. An approximate segmentation of the anomaly is then obtained via the thresholded difference between the original and the pseudo-healthy image. Recent DDPM-based approaches often perform the reconstruction in a multi-stage fashion: First, a rough map anomaly map is computed, and then the highlighted areas are in-painted with healthy-looking tissue~\cite{autoddpm,pinaya2022fast}. While these multi-stage approaches may boost performance, the additional steps also add to the complexity and execution time of the UAD pipeline. 

Since anomaly maps are typically continuous, a key question in the UAD literature is how to set the threshold to binarize a segmentation, and consequently, whether a sample can be considered anomalous~\cite{baur2021autoencoders}. In practice, many previous approaches directly tune the threshold for lesion detection on labeled data from the target domain, which overestimates detection performance and limits the approaches' generalizability to other domains~\cite{pinaya2022fast,autoddpm,dae,normative-ascent-elbo,baur2021autoencoders,bercea2023reversing}. Conceptually, such approaches cannot be considered truly unsupervised since they require labeled data for determining the optimal operating point. 

Knowledge about the target domain can also be introduced via the selection of the modalities on which the algorithm is trained~\cite{anoddpm,autoddpm,normative-ascent-elbo,pinaya2022fast}. For example, tumors are easiest to detect on T2-weighted or FLAIR images, while chronic post-stroke changes are best detected in T1-weighted images. Evaluating only the modality on which the target anomaly is easiest to identify, again, overestimates detection performance in a truly unsupervised scenario. Most prior work is currently evaluated with a single image modality (\eg~\cite{anoddpm,autoddpm,pinaya2022fast,normative-ascent-elbo}), which makes it unclear how well these methods would generalize to the clinically relevant multi-modal scenario.  

Lastly, a number of techniques use implicit assumptions about the appearance of the target anomalies. For instance, a recent approach based on denoising autoencoders (DAEs) employs coarse noise during training that mimics the structure of large tumors on which the approach is evaluated~\cite{dae}. However, as we show, the approach does not generalize to smaller multiple sclerosis (MS) lesions. 

In this paper, we propose Aggregated Normative Diffusion models \texttt{ANDi}, a novel DDPM-based unsupervised anomaly detection algorithm that requires no inpainting stage, is simple to use, and achieves state-of-the-art performance on diverse multi-modal neuroimaging datasets in a realistic multi-modal MR imaging setting. Anomaly maps are obtained by directly aggregating deviations between the actual and the learned transitions in the denoising process (see Fig.\,\ref{fig:teaser}). Key to the performance of our proposed method is the use of multi-resolution pyramidal (Gaussian) noise for training the DDPM. This substantially boosts performance and allows the method to generalize to different types of pathologies. We also introduce a novel mechanism for aggregating deviations in the diffusion process based on the geometric mean that leads to further improvements. Crucially, \texttt{ANDi} works with no a priori modality selection and does not require threshold tuning on labeled data.

Our contributions are as follows:
\begin{enumerate}

    \item We demonstrate that current state-of-the-art UAD methods are not robust to varying lesion types in the clinically relevant multi-modal MR scenario. 
    \item We propose \texttt{ANDi}, a simple and fast DDPM-based UAD method based on aggregating deviations between the actual and the learned denoising transition, a novel pyramidal noise, and a novel aggregation mechanism. 
    \item We introduce a novel domain-agnostic thresholding strategy based on Yen thresholding~\cite{yen1995new} that can be applied to any method and allows evaluation of segmentation accuracy without domain-specific threshold tuning. 
    \item We show that our method consistently outperforms related work on three diverse multi-modal brain anomaly datasets, both in a domain-agnostic evaluation and in terms of the commonly used ``ceil'' Dice score. 
\end{enumerate}

\section{Related Work}

The first methods tackling anomaly detection in medical images originated in the field of neuroimaging. The earliest approaches rely on building a population-based brain atlas reflecting an ``average'' normal subject and registering potentially anomalous images to that atlas. Unmatchable areas are then used to identify potential pathologies~\cite{cabezas2011review,van2001automated,prastawa2004brain}. A limitation of these approaches is that the registration step may fail in the presence of anomalies. While this issue can be partially mitigated with registration frameworks that are more robust to anomalies (e.g.\cite{liu2014low}), handling anomalies that deviate significantly from the atlas remains a problem. Moreover, the creation of an atlas can be subjective and may not represent all patient populations equally. 

More recently, Meissen \etal\cite{meissen2021challenging} proposed a simple semi-supervised baseline approach for tumor and white matter lesion segmentation in which the pre-processed FLAIR modality is simply thresholded in image space. The optimal threshold is determined with a labeled validation set. This naive approach works surprisingly well for tumors and white matter lesion detection, as they appear bright compared to the surrounding tissue in FLAIR images. However, this approach is data-specific and will not generalize to more complex types of anomalies. In this paper, we use this approach as a semi-supervised baseline to our proposed fully unsupervised method. 

\subsection{Deep Learning-based UAD in Medical Imaging}

The emergence of deep learning-powered generative models such as VAEs~\cite{kingma2013auto}, or GANs~\cite{goodfellow2014generative} has spurred a new generation of UAD methods. In one of the earliest approaches, Sato \etal~\cite{sato2018primitive} use an AE framework to detect anomalies in computed tomography images of the head. This work was followed by a series of contributions based on AEs~\cite{baur2020bayesian}, VAEs~\cite{chen2018unsupervised,normative-ascent-elbo,bercea2023generalizing,bercea2023reversing}, and GANs~\cite{f-anogan,schlegl2017unsupervised}. For the most part, these techniques rely on generating a pseudo-healthy image from a potentially anomalous image. 

Kascenas \etal~\cite{dae} proposed to train a denoising autoencoder (DAE) on multi-modal brain MR images with a custom ``coarse'' noise. While simple, this approach reached state-of-the-art performance and, to date, was not conclusively outperformed by newer approaches. Indeed, in more recent work, Kascenas \etal~\cite{KASCENAS2023102963} showed that the technique performs competitively when compared to recent DDPM-based approaches. Moreover, this technique is one of the few to consider the realistic multi-modal scenario where the network is trained and evaluated with four diagnostically important MR sequences per patient.

Recently, DDPMs~\cite{ddpm} have been shown to reach unprecedented performance for modeling high-dimensional probability distributions, and a number of recent works build on this approach for UAD~\cite{anoddpm,autoddpm,pinaya2022fast}.

Similar to our proposed work, Pinaya \etal~\cite{pinaya2022fast} aggregate the Euclidian distances between the ground truth backwards transition and the denoising step of the diffusion model. In contrast to our work, these aggregated distances are used to obtain a preliminary anomaly mask, which is then inpainted via another pass through the DDPM model. Interestingly, Kascenas \etal~\cite{KASCENAS2023102963} showed that omitting the inpainting step leads to superior performance, however, the method still does not outperform the simpler DAE~\cite{dae}. 

In closely related work, Wyatt \etal~\cite{anoddpm} introduce AnoDDPM, a pseudo-healthy generation method based on DDPMs. Similar to our work, the authors employed a multi-scale noise in the training of the DDPM in order to capture anomalies at different scales. Specifically, the authors use multi-scale simplex noise~\cite{perlin2002improving}. However, generating simplex noise incurs a comparatively large computational cost. In contrast, our proposed pyramidal Gaussian noise is much faster and easier to generate, and empirically leads to a better performance within our framework. 

Very recently, Bercea \etal~\cite{autoddpm} proposed a DDPM-based UAD framework based also on pseudo-healthy image generation. The pipeline uses the DDPM to produce an initial anomaly likelihood map, which is then in-painted in an 
iterative process. The method was shown to outperform the earlier AnoDDPM~\cite{anoddpm} on stroke lesions. However, as we will show in our experiments, the method is currently not outperforming the simple DAE approach ~\cite{dae}.

\section{Denoising Diffusion Probabilistic Models}

\begin{algorithm}[t]
\caption{\texttt{ANDi} for unsupervised anomaly detection.}
\textbf{Input}: Initial volume $v \in \mathbb{R}^{H \times W \times D \times C}$, $T_l$, $T_u$, noise distribution $N_\epsilon$, thresholding $h:a \rightarrow s$.
\begin{algorithmic}[1]
\State $a_m \gets \{\}$ \Comment{empty multi-modal anomaly map}
\For{each $x_0\in\mathbb{R}^{H \times W \times C}$ from $v$}
\For{each $t \in \{T_l, \ldots, T_u \}$}
    \State $\epsilon_t\sim N_\epsilon$ \Comment{Sample noise}
    \State $x_t \gets \sqrt{\bar{\alpha}_t} \cdot x_0 + \sqrt{1-\bar{\alpha}_t} \cdot \epsilon_t$\Comment{Noise the image}
    \State $\tilde{x}_{t-1} \gets \mu_q(x_t, x_0) $ \Comment{Expected transition}
    \State $\hat{x}_{t-1} \gets \mu_\theta(x_t, t)$ \Comment{Predicted transition}
    \State $d_t \gets \left( \tilde{x}_{t-1} - \hat{x}_{t-1}\right)^2$ \Comment{Get deviation (\cref{eq:dist})}
\EndFor
\State $i_m = A_{\text{gm}}(\{d_t\})$\Comment{Aggregate per modality (\cref{eq:geometric-mean}})
\State $a_m\oplus i_m$ \Comment{Concatenate volume slice}
\EndFor
\State $a = \max_m a_m$ \Comment{Max over modalities (\cref{sec:anomaly-to-segmentation})}
\State $ s \gets h(a)$ \Comment{Thresholding (\cref{sec:anomaly-to-segmentation})}
\end{algorithmic}
\textbf{Output}: Binarized anomaly map $s\in \{0,1\}^{H \times W \times D}$.
\label{alg:main}
\end{algorithm}

\label{back:dpm}
DDPMs \cite{ddpm} are a class of generative models that aim to approximate a data distribution $q(x_0)$ given some training data from that distribution. A DDPM consists of a fixed forward diffusion (noising) process and a learned reverse denoising process. The forward process gradually adds Gaussian noise $\epsilon$ to an input image $x_0$ in a fixed forward process ($x_{t-1}\xrightarrow{\text{noise}} x_t$), transforming it into Gaussian noise over $T$ steps. A neural network then learns to reverse this process iteratively ($x_{t}\xrightarrow{\text{denoise}} x_{t-1}$) to be able to generate samples from the original data distribution.

The forward noise step from the original image $x_0$ to the image $x_t$ at timestep $t$ is given by the Gaussian transition 
\begin{equation}
\label{eq:forward-transition}
 q(x_t\vert x_0) = \mathcal{N}\left( \sqrt{\bar{\alpha}_t}x_0, (1-\bar{\alpha}_t)I \right),
\end{equation}
where $\alpha_1\ldots \alpha_T$ are scaling factors that follow a fixed schedule, and $\bar{\alpha}_t=\prod_{k=1}^t \alpha_k$. To generate an image from noise, we require sampling from the reverse transition $q(x_{t-1}\vert x_t)$, which is intractable. However, as described in~\cite{ddpm}, we can apply Bayes rule to compute the closed-form expression for the reverse transitions \textit{conditioned on} $x_0$ as
$q\left(x_{t-1}\vert x_t, x_0\right)=\mathcal{N}\left(\mu_q(x_t, x_0), \Sigma_q(t)\right)$, where 
\begin{align}
\label{eq:mean}
\hspace{-0.65em}\mu_q(x_t, x_0)=\frac{\sqrt{\alpha_t}(1-\bar{\alpha}_{t-1})}{1-\bar{\alpha}_t}x_t + \frac{\sqrt{\bar{\alpha}_{t-1}}(1-\alpha_t)}{1-\bar{\alpha}_t}x_0.%
\end{align}
We train the model to approximate the true denoising distribution $q(x_{t-1}\vert x_t)$ by the variational posterior 
\begin{equation}
\label{eq:forward-transitions}
    p_\theta(x_{t-1}\vert x_t)=\mathcal{N}\left(\mu_\theta(x_t, t), \Sigma_\theta(x_t, t)\right),
\end{equation}
 for each step $t$, where a neural network is used to estimate the mean $\mu_\theta(x_t, t)$ while the variance $\Sigma_\theta(x_t, t)$ is set to $\Sigma_q(t)=\frac{(1-\alpha_t)(1-\bar{\alpha}_{t-1})}{1-\bar{\alpha}_t}I$.%

The model parameters $\theta$ are learned by maximizing the Evidence Lower Bound on the log-likelihood of the joint distribution $q(x_0\ldots x_T)$, which can be decomposed into a sum of KL divergence terms of the form $D_{KL}\left [q(x_{t-1}\vert x_t,x_0) \Vert p_\theta(x_{t-1}\vert x_t)\right]$ for each denoising step $t$. The training objective can thus be written as
\begin{align}
\label{eq:objective}
    \hspace{-0.6em}\underset{\theta}{\arg\min}~\mathbb{E}_{x_{0\ldots T}}\sum_{t=1}^T \Sigma^{-1}_q(t)\Vert \mu_\theta(x_t, t) - \mu_q(x_t, x_0) \Vert^2.
\end{align}
In practice, a simpler training objective is used, where the scaling terms $\Sigma^{-1}_q(t)$ are omitted and the predicted and target means are reparameterized in terms of the noise $\epsilon$, i.e.
\begin{align}
\label{eq:simpler_objective}
    \underset{\theta}{\arg\min}~\mathbb{E}_{x_0,\epsilon_0}\sum_{t=1}^T\Vert \epsilon_\theta(x_t, t) - \epsilon_0 \Vert^2, \text{with}
\end{align}%
$\epsilon_0$ being the noise added to $x_0$ to get $x_t$.
Regardless of reparameterization, this objective essentially consists of a sum of the discrepancies between the predicted and target means. In the following section, we demonstrate how the form of the optimization objective can be used to create anomaly maps by aggregating these discrepancies.

\section{Aggregated Normative Diffusion \texttt{ANDi}}
Given a potentially anomalous multi-modal volume $v \in \mathbb{R}^{H \times W \times D \times C}$, the aim of UAD is to obtain a pixel-wise anomaly map $a$ indicating the likelihood of an anomaly being present, where $C$ is the number of image modalities, and $H,W,D$ are the image dimensions. An anomaly is anything that deviates from the anatomy of healthy subjects. In the multi-modal setting, we initially obtain an anomaly map $a_m$ for each channel, which then needs to be reduced to a pixel-wise single-channel anomaly map $a$. From the anomaly map $a$, a binary anomaly segmentation mask $s\in \{0,1\}^{H \times W \times D}$ can then be generated using a thresholding mechanism. In practice, $s$ could, for example, be the segmentation of a tumor or an MS lesion.

In the following, we first describe how we train a DDPM to model the normative reference distribution (\cref{sec:ddpm-training}), and how that model can be used to obtain anomaly maps via aggregation of differences between the ground truth backwards transitions and the denoising step (\cref{sec:aggregation}). In \cref{sec:geometric-mean}, we describe a novel approach for aggregating those differences, which leads to more robust anomaly detection, and in \cref{sec:pyramidal_noise} we describe a pyramidal noise type that can be used in DDPM training and further improves performance. Lastly, in \cref{sec:anomaly-to-segmentation} we describe how multi-channel anomaly maps $a_m$ are reduced to single-channel anomaly maps $a$, and how we obtain final anomaly segmentations from them. An overview of our final anomaly detection algorithm can be found in \cref{alg:main}.

\subsection{Learning the Normative Distribution using DDPMs}
\label{sec:ddpm-training}

Similar to recent work \cite{anoddpm,pinaya2022fast,autoddpm}, we use a DDPM~\cite{ddpm} to learn the \textit{normative distribution} $q_{norm}(x)$ of normal image \emph{slices} $x\in \mathbb{R}^{H \times W \times C}$ given a training set of healthy image slices $\{ x^{(0)}, \ldots, x^{(M)} \}$.

We empirically observed that training with standard Gaussian noise transitions given in \cref{eq:forward-transitions} do not allow our final model to capture anomalies at different scales. Instead, we train our DDPM with pyramidal Gaussian noise, which will be introduced in \cref{sec:pyramidal_noise}.

\subsection{Generating Anomaly Maps via Aggregation of Denoising Errors}
\label{sec:aggregation}

After training, we use the learned denoising step along with the ground truth backwards process to obtain an anomaly map. The denoising step can be thought of as \emph{normative diffusion} as it is taking one step towards the normative distribution, while the ground truth backwards transition is going back towards the anomalous image. The discrepancy between the two is therefore indicative of potential anomalies.

Specifically, for a potentially anomalous multi-modal input image $x_0$, we first obtain the noised images $x_{t}$ for all $t \in [T_l, T_u]$ using the forward process (see \cref{eq:forward-transition}). We use $T_l = 75$, $T_u = 200$ for all experiments. Next, for each time step in that range, we obtain the KL-divergence term (\cref{eq:objective}) which amounts to the Euclidean distance $d_t$ between the ground truth backwards transition $\tilde{x}_t = \mu_q(x_t, x_0)$ and the learned denoising step $\hat{x}_t = \mu_\theta(x_t, t)$, 
\begin{equation}
    \label{eq:dist}
    d_t = \left(\mu_q(x_t, x_0) - \mu_\theta(x_t, t)\right)^2.
\end{equation}

Importantly, each $d_t$ term can be computed independently. Thus, the entire computation can be parallelized across time steps, which has the potential for significant speedups compared to DDPM approaches performing pseudo-healthy image generation typically requiring step-by-step denoising (\eg~\cite{anoddpm,autoddpm}). 

In order, to obtain the initial multi-channel anomaly maps $a_m$, we aggregate the individual $\{d_t\}$ with $t \in [T_l, T_u ] $ using some aggregation function $A$. In this work, we explore a simple aggregation via the arithmetic mean over time steps,
\begin{equation}
    \label{eq:arithmetic-mean}
    A_{\text{am}}(\{d_t\}) = \frac{1}{T_u-T_l} \sum_{t={T_l}}^{T_u} d_t,
\end{equation}
as well as a novel aggregation mechanism based on the geometric mean which we will describe next. 

\subsection{Geometric Mean for Error Aggregation}
\label{sec:geometric-mean}

The arithmetic mean in \cref{eq:arithmetic-mean} can be sensitive to outliers when aggregating across timesteps. To address this, we propose to use the more robust \emph{geometric mean} for aggregation:
\begin{equation}
    \label{eq:geometric-mean}
    A_{\text{gm}}(\{d_t\}) = \exp \left( \frac{1}{T_u-T_l} \sum_{t={T_l}}^{T_u} \log d_t \right).
\end{equation}
This approach dampens the effect of extreme values across timesteps. We empirically show that the geometric mean leads to better results for anomaly detection compared to the arithmetic mean used in \cite{pinaya2022fast}.

\begin{figure}[t]
    \centering
    \includegraphics[width=\linewidth]{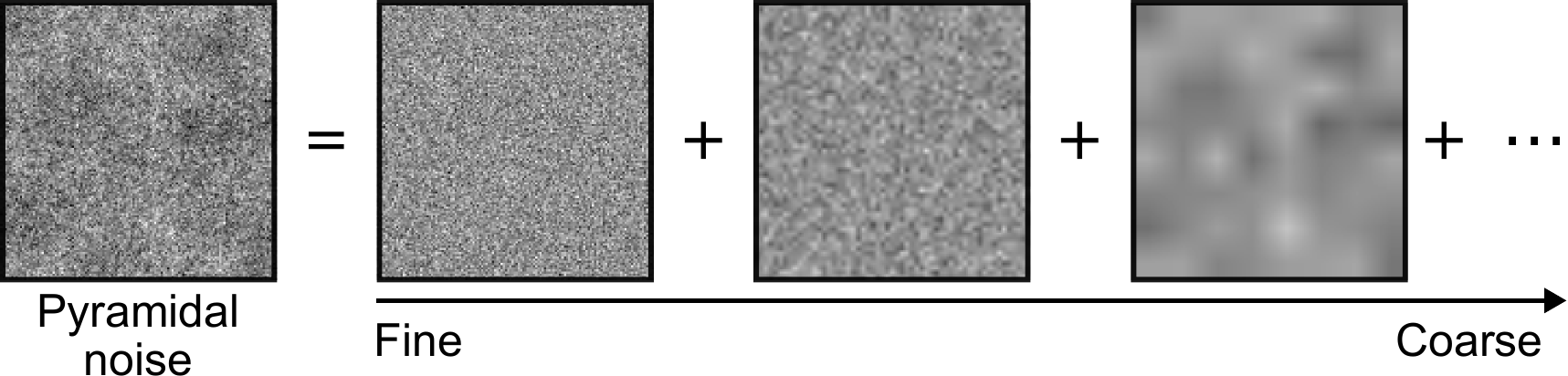}
    \caption{An illustration of pyramidal Gaussian noise.}
    \label{fig:pyramidal-noise}
\end{figure}

\subsection{Robust Anomaly Detection via Pyramidal Noise}
\label{sec:pyramidal_noise}

In order to make the denoising transitions more sensitive to anomalies at different scales (e.g. different lesion sizes), we propose to train our model using a multi-resolution pyramidal Gaussian noise in which noise is sampled at increasing image resolutions, bilinearly upsampled to the original image resolution, and then added. This type of noise was previously explored for improved image generation in the following blog post~\cite{multires_wandb}. An illustration of pyramidal noise is shown in \cref{fig:pyramidal-noise}. 

Mathematically, we define pyramidal noise as
\begin{equation}
    \label{eq:pyramidal_noise}
    \epsilon = \sum\limits_{i=1}^{10} c^i \cdot U(\epsilon^{(i)}; H, W),
\end{equation}
where $c$ is a constant scaling factor ($c=0.8$ for all experiments), $\epsilon^{(i)} \sim \mathcal{N}(0, I_{h_i\times w_i\times C})$ is independent standard Gaussian noise of size $h_i\times w_i \times C$, and $U$ is a bilinear upsampling operation that upscales the images to size $H\times W \times C$. The Gaussian noise height and width $h_i,w_i$ are calculated by successively dividing the base width and height in half for each level. Following \cite{multires_wandb}, a slight jitter is added to these values at each level, resulting in the following equations 
\begin{align}
    h_i =& \lceil H/(r_i)^{i-1}\rceil \\\nonumber
    w_i =& \lceil W/(r_i)^{i-1}\rceil \\\nonumber
         & \text{with} \ \ r_i = 2+\text{Unif}(0,2).
\end{align}
Finally, we normalize the pyramidal noise to unit variance. 

Similar to the multi-scale simplex noise previously proposed in AnoDDPM~\cite{anoddpm}, the pyramid noise allows the learned backwards transitions to denoise at multiple spatial scales and thus increases robustness across different anomaly sizes. However, in contrast to multi-scale simplex noise, pyramidal noise is very simple to generate and incurs much smaller computational costs.
In our experiments, we empirically show that we obtain superior performance compared to the AnoDDPM which uses simplex noise. 

The pyramidal noise may be used for both the training (\cref{sec:ddpm-training}) and anomaly map generation (\cref{sec:aggregation}). However, we found that training with pyramidal noise, but generating anomaly maps with normal Gaussian noise consistently improved performance (see also ablation study in \cref{tab:ablation}). This can be explained by the following intuition: During training, the denoising function $\mu_\theta(x_t, t)$ in \cref{eq:forward-transition} learns to remove pyramidal noise. During evaluation, we only noise the images with Gaussian noise. This means that the denoising function tries to remove the simple Gaussian noise plus the additional lower-frequency components. Since anomalies may resemble lower-frequency noise, the backwards transition is more likely to remove them. This is similar to the intuition of the DAE approach in \cite{dae}, where the denoising autoencoder is trained with coarse noise and evaluated with no noise.

\subsection{From Anomaly Scores to Segmentation Masks}
\label{sec:anomaly-to-segmentation}

Given a potentially anomalous multi-channel input volume $v$, we first slice it into axial multi-channel slices. Then, for each slice, we obtain the multi-channel anomaly maps, which are then stacked back into an anomaly volume $a_m$. In order to reduce the channel-wise anomaly map to a pixel-wise anomaly map $a$, prior work has used the arithmetic mean across the channel dimension~\cite{dae}. In this work, we additionally investigated a $max$ operation across channels. We found that this led to slightly better performance for all examined methods, including the DAE approach in~\cite{dae}, and thus used the $max$ operation for all experiments.

Following related work (\eg~\cite{baur2021autoencoders}), we then apply a 3-dimensional median filtering to the anomaly maps. We explore kernel sizes of $3\times 3 \times 3$, and $5\times 5 \times 5$. We also apply median filtering to all baselines and find that it improves all examined techniques. 
            
In order to obtain a binarized segmentation mask or a decision about the abnormality of a sample, thresholding must be performed. In reality, it is non-trivial to find a good threshold without making use of any labeled data. Many prior UAD works indeed do use labels to obtain the binarized segmentation masks~\cite{pinaya2022fast,autoddpm,dae,normative-ascent-elbo,baur2021autoencoders,bercea2023reversing}. Additionally, the ceil Dice is a widely used metric in UAD. This metric is derived by creating a segmentation mask $s$ using an optimal threshold that is determined based on the test data, and then comparing this mask to the ground truth segmentation. While this approach provides an upper limit of performance, it does not reflect the true unsupervised performance in scenarios where labeled data for tuning is unavailable. 

As an additional domain-agnostic thresholding strategy, we propose to automatically threshold the anomaly maps using the Yen technique~\cite{yen1995new}. This approach does not require access to the test dataset and allows to obtain a unique threshold for every subject. In that sense, it allows evaluation in a truly unsupervised setting with no knowledge of the target application. We observed that Yen thresholding leads to consistent underestimation of the anomaly size. Therefore, we additionally post-process the resulting segmentation with a 3D dilation step. We additionally report the Dice score obtained in this manner ($\text{Dice}_{\text{Yen}}$) for all methods examined in this paper.

\begin{figure*}[ht]
    \centering
    \includegraphics[width=\linewidth]{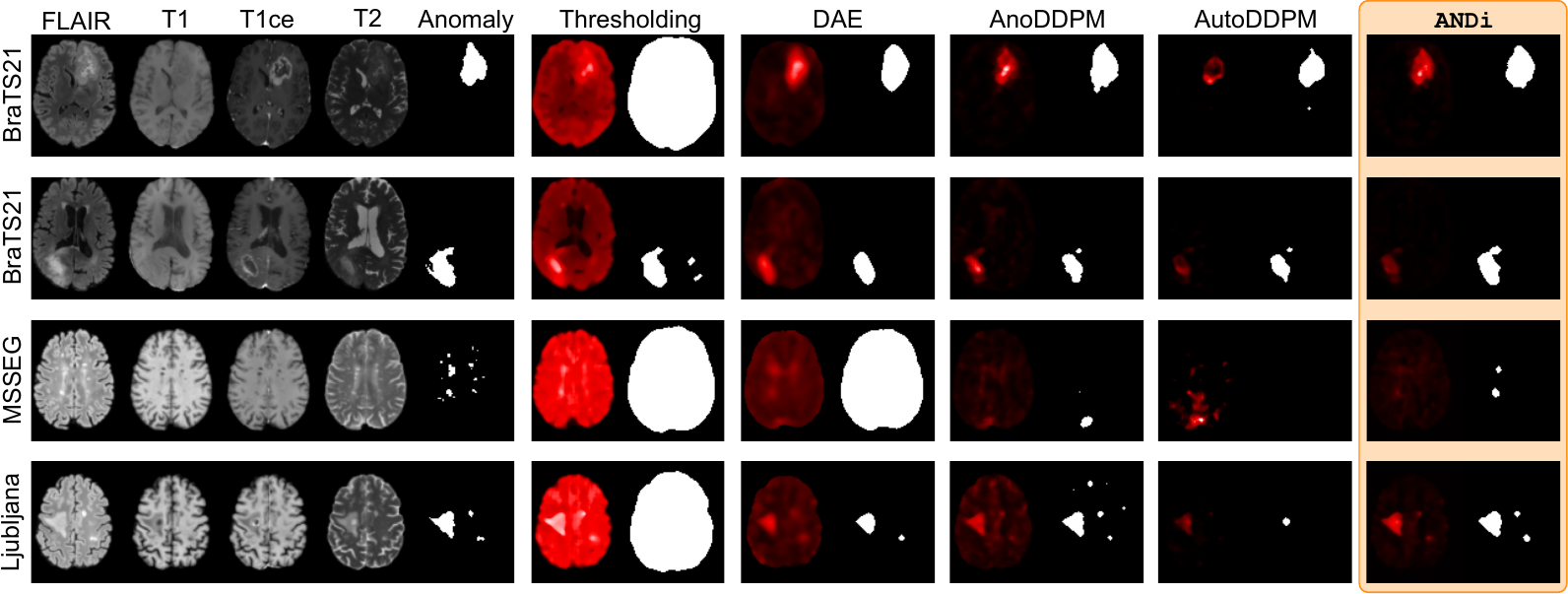}
    \caption{Comparative analysis of inferred anomalies and binarized segmentations using Yen thresholding\cite{yen1995new} across baseline methods and \texttt{ANDi}. These anomaly maps, enhanced through modality max-pooling and median filtering (kernel size 5), highlight the limitations of calibration-dependent FLAIR thresholding\cite{meissen2021challenging} and DAE's\cite{dae} inefficiency with small lesions. Among methods based on diffusion models, both AnoDDPM\cite{anoddpm} and \texttt{ANDi} demonstrate proficiency in isolating anomalies of varying sizes, with \texttt{ANDi} showing superior performance in unsupervised thresholding. The anomaly map brightness is volume-scaled, potentially appearing darker in individual slices.}
    \label{fig:qualitative}
\end{figure*}
\section{Experiments \& Results}
\subsection{Data}
We evaluated \texttt{ANDi} on data from three multi-modal brain MR datasets. Namely, we used images from the brain tumor segmentation (BraTS'21) challenge \cite{brats21} as well as the MSSEG~\cite{commowick2018objective} and Ljubljana MS lesion datasets, which are both part of the 2021 Shifts challenge \cite{shifts2021}. Each subject in these datasets comprises the following four MR scans: native T1-weighted, post-contrast
T1 gadolinium weighted (T1Gd), T2-weighted (T2), and Fluid Attenuated Inversion Recovery (FLAIR).

\subsection{Preprocessing}

All datasets have been skull-stripped and resampled to a unified resolution. We removed one subject from the Ljubljana dataset where the skull-stripping has failed. The volumes and segmentation masks from the Shifts data were coregistered to the SRI24 atlas \cite{rohlfing2010sri24}. After registration, the segmentation masks were binarized using a threshold of 0.5. Afterward, we applied pairwise histogram matching \cite{nyul2000new} between all subjects from the Shifts datasets to a randomly chosen subject of the BraTS'21 dataset in order to reduce domain shifts resulting from different acquisition protocols.

All volumes were individually divided by the 99th percentile foreground voxel intensity for each subject and modality. The axial slices were downsampled to a size of 128x128 using bilinear downsampling with anti-aliasing. The ground truth segmentation masks were downsampled to the same size using nearest neighbor interpolation.

\subsection{Experiments}

Similar to \cite{dae}, we used the healthy slices in the BraTS'21 dataset to train all our models. The BraTS'21 dataset is comparatively large consisting of MR images of 1251 subjects, and contains considerable diversity in contrasts. Using the data splits previously proposed by \cite{dae}, we divided the dataset into 938 subjects for the training, 62 subjects for the validation, and 251 subjects for the testing. In the training split, we then identified all slices that did not contain any tumors using the ground truth segmentations. The training slices that do contain tumor tissue are discarded. The healthy BraTS'21 training slices were then used to train our proposed as well as all baseline methods. 

We evaluated the different UAD algorithms examined in this study on all of the test subjects of the BraTS'21 dataset, the 52 subjects of the MSSEG dataset, and the 24 subjects of the Ljubljana dataset. 

\begin{table*}[ht]
\centering
\caption{Anomaly detection performance measured in AUPRC ($\uparrow$), $\lceil \text{Dice} \rceil$ ($\uparrow$) and $\text{Dice}_{\text{Yen}}$ ($\uparrow$) on the test dataset of BraTS'21 (251 subjects) and Shifts Challenge (MSSEG -- 52 subjects, Ljubljana -- 24 subjects). We highlight the \fcolorbox{white}{tabfirst}{\textbf{first}}, \fcolorbox{white}{tabsecond}{second}, and \fcolorbox{white}{tabthird}{third} best-performing unsupervised methods for each metric. We exclude the semi-supervised thresholding technique from the ranking as it requires labeled data for segmentation threshold tuning, and relies on an a priori modality selection.}
\label{tab:main-results}
\small
\begin{tabular}{llccccccccc}
 & & \multicolumn{3}{c}{BraTS21} & \multicolumn{3}{c}{Shifts -- MSSEG}& \multicolumn{3}{c}{Shifts -- Ljubljana }\\ 
\cmidrule(lr){3-5} \cmidrule(lr){6-8} \cmidrule(lr){9-11}
\multicolumn{2}{c}{Method} & AUPRC& $\lceil \text{Dice} \rceil$ &$\text{Dice}_{\text{Yen}}$& AUPRC& $\lceil \text{Dice} \rceil$ &$\text{Dice}_{\text{Yen}}$& AUPRC& $\lceil \text{Dice} \rceil$ &$\text{Dice}_{\text{Yen}}$\\
\toprule
\multirow{3}{*}{\rotatebox[origin=c]{90}{\textit{\footnotesize Semi-sup.}}}~~Thresholding (FLAIR)~\cite{meissen2021challenging} &&0.741&0.674&0.106&0.430&{0.331}&0.018&{0.425}&0.345&0.027\\
~~~~~~~~~~w/ MF size 3&&0.795&0.712&0.106&{0.465}&0.303&0.018&0.403&0.302&0.027\\
~~~~~~~~~~w/ MF size 5&&0.818&0.726&0.106&0.386&0.231&0.018&0.304&0.226&0.023\\
\addlinespace[0.15em]
\midrule
\multirow{13}{*}{\rotatebox[origin=c]{90}{\footnotesize\textit{Unsupervised}}}~~DAE \cite{dae} &&0.810&0.724&0.700&0.022&0.117&0.037&0.138&0.140&0.077\\
~~~~~~+ Modality max pooling&&0.817&0.733&0.699&0.049&0.126&0.043&0.059&0.100&0.041\\
~~~~~~~~~~w/ MF size 3&&\cellcolor{tabthird}0.835&\cellcolor{tabthird}0.752&\cellcolor{tabthird}0.740& 0.048&0.126&0.054& 0.066& 0.099&0.069\\
~~~~~~~~~~w/ MF size 5&&\cellcolor{tabsecond}0.845&\cellcolor{tabsecond}0.763&\cellcolor{tabsecond}0.759&0.041&0.120&0.058&0.070&0.096&0.070\\
~~~~~AnoDDPM\cite{anoddpm}&&0.600&0.532&0.488&0.104&0.140&0.150&0.070&0.105&0.062\\
~~~~~~~~~~w/ MF size 3&&0.731&0.650&0.609&\cellcolor{tabthird}0.187&0.194&0.214&0.137&0.139&0.128\\
~~~~~~~~~~w/ MF size 5&&0.769&0.687&0.657&0.156&0.169&0.167&0.152&0.131&0.129\\
~~~~~AutoDDPM \cite{autoddpm}~~~~&&0.543&0.511&0.437&0.084&0.144&0.152&0.103&0.192&0.133\\
~~~~~~~~~~w/ MF size 3&&0.686&0.630&0.558&0.134&0.178&0.202&0.128&0.210&0.213\\
~~~~~~~~~~w/ MF size 5&&0.731&0.674&0.606&0.108&0.127&0.150&0.076&0.155&0.155\\
~~~~ \texttt{ANDi} (\textit{ours}) &&0.654&0.587&0.388&\cellcolor{tabsecond}0.219&\cellcolor{tabsecond}0.274&\cellcolor{tabsecond}0.290&\cellcolor{tabsecond}0.325&\cellcolor{tabthird}0.331&\cellcolor{tabthird}0.228\\
~~~~~~~~~~w/ MF size 3 &&0.829&0.747&0.701&\cellcolor{tabfirst}\textbf{0.321}&\cellcolor{tabfirst}\textbf{0.311}&\cellcolor{tabfirst}\textbf{0.380}&\cellcolor{tabfirst}\textbf{0.422}&\cellcolor{tabfirst}\textbf{0.347}&\cellcolor{tabfirst}\textbf{0.356}\\
~~~~~~~~~~w/ MF size 5 &&\cellcolor{tabfirst}\textbf{0.858}&\cellcolor{tabfirst}\textbf{0.781}&\cellcolor{tabfirst}\textbf{0.766}&0.174&\cellcolor{tabthird}0.201&\cellcolor{tabthird}0.240&\cellcolor{tabthird}0.293&\cellcolor{tabthird}0.242&\cellcolor{tabsecond}0.253\\
\bottomrule
\end{tabular}
\vspace{-0.2em}
\end{table*}

\subsection{Metrics}

For all methods, we report the following three metrics: Area under the precision-recall curve (AUPRC), ceil Dice ($\lceil \text{Dice} \rceil$), and the fully unsupervised $\text{Dice}_{\text{Yen}}$ values, where no prior domain knowledge is used to tune the segmentation threshold (see \cref{sec:anomaly-to-segmentation}). To calculate the AUPRC, the anomaly scores are normalized to be in the range of $[0, 1]$. AUPRC and $\text{Dice}_{\text{Yen}}$ evaluate the performance of the method without assumptions about the target domain, while $\lceil \text{Dice} \rceil$ uses the test dataset to determine an optimal threshold and gives an upper-bound on the achievable performance. 

\subsection{Baselines} 

In addition to our proposed \texttt{ANDi} approach, we also evaluate 3 state-of-the-art UAD baselines: DAEs~\cite{dae}, AnoDDPM~\cite{anoddpm}, and AutoDDPM \cite{autoddpm}. As an additional baseline, we consider the semi-supervised thresholding technique by Meissen \etal~\cite{meissen2021challenging} where we used the FLAIR modality. 

Out of the UAD baselines, only the DAE~\cite{dae} was specifically designed for multi-modal data. The original method uses the arithmetic mean to combine the anomaly maps for the different input modalities. We additionally report results with our proposed max modality pooling (see \cref{sec:anomaly-to-segmentation}), showing that it improves results in 2 out of 3 test datasets. Since the other methods were all proposed for a single modality scenario, we by default extend them by the max modality pooling operation to make them multi-modality compatible.  
AutoDDPM \cite{autoddpm} requires a masking threshold for their first mask calculation that was obtained using the 95th percentile of the residuals. In our experiments we found that the 99th percentile worked better on the validation dataset of BraTS'21, and we used this threshold for all our analyses. 

Lastly, we report all results with median filtering with a $3\times 3 \times 3$ (MF size 3) and a $5\times 5 \times 5$ (MF size 5) kernel, as we found the results to be particularly sensitive to this hyperparameter choice.

\subsection{Implementation Details}
In line with prior work, we used a U-Net to implement our DDPM model~\cite{anoddpm,autoddpm,pinaya2022fast,ddpm} as well as all DDPM-based baselines in this paper. We trained the models with the AdamW optimizer with standard hyperparameters. The learning rate is changed per step and follows a linear warmup with a cosine decay. A description of all hyperparamters can be found in \cref{appendix:a}. 

Unlike usual DDPM implementations, we did not clip the predictions at each timestep, as clipping may remove important outlier values from the anomaly maps. We have found minor improvements when neglecting the clipping for a Gaussian and pyramidal noise schedule.

\section{Results}

Our main results for all baselines on the three datasets are reported in \cref{tab:main-results} with qualitative results shown in \cref{fig:qualitative}. We find that our proposed \texttt{ANDi} consistently and, on some datasets, substantially outperforms all unsupervised baselines. Importantly, \texttt{ANDi} is the first UAD algorithm to surpass the strong baseline of DAEs~\cite{dae}, which had never been outperformed on the BraTS'21 dataset. On the more challenging MS lesion datasets (MSSEG \& Ljubliana), all methods perform significantly worse. However, \texttt{ANDi} clearly outperforms the baselines with and without median filtering. On the AUPRC metric, the best-performing configuration of our method outperforms the next best method by 72\% on the MSSEG dataset and by 178\% on the Ljubliana dataset. While the strongest DAE~\cite{dae} configuration performs very closely to our strongest configuration on BraTS'21, it performs comparatively poorly on the Shifts datasets. This suggests that the DAE approach may be specifically tailored to brain tumor detection due to its use of coarse noise during training, which resembles large anomalies. 

For small lesions, we find that MF size 5 degrades the performance of all unsupervised methods since the smoothing operation leads to small lesions being discarded from the anomaly map. We also observe that the domain-agnostic Yen thresholding achieves Dice scores similar to $\lceil \text{Dice} \rceil$ when median filtering is applied. This implies that the Yen thresholding yields results close to the theoretical Dice upper bound when using the test set to tune the threshold.

We observe that for BraTS'21 our method and the DAE approach~\cite{dae} outperform the semi-supervised thresholding baseline~\cite{meissen2021challenging}. We emphasize that thresholding is not directly comparable to the unsupervised methods because it can, by definition, only detect hyperintense lesions, and requires labeled data to determine a threshold for binary segmentation. The fact that it would not perform well in a realistic scenario is reflected by the low $\text{Dice}_{\text{Yen}}$ values. 

In order to better understand the contribution of each of our proposed novelties, we also performed an ablation study on the BraTS'21 validation shown in \cref{tab:ablation}. We started with the \texttt{ANDi}$_\text{base}$ algorithm, which is a DDPM trained with standard Gaussian noise, and anomaly map generation via aggregation of KL-terms using the arithmetic mean as described in \cref{sec:aggregation}. We then sequentially added the modality max pooling (\cref{sec:anomaly-to-segmentation}), the geometric mean aggregation (\cref{sec:geometric-mean}), training and testing with pyramidal noise (\cref{sec:pyramidal_noise}), median filtering with kernel size 5 (\cref{sec:anomaly-to-segmentation}), and finally training with pyramidal noise but evaluating with standard Gaussian noise (\cref{sec:pyramidal_noise}). We find that each of these contributions is crucial for achieving the final performance.

\begin{table}
\centering
\caption{Ablation study of our individual contributions on the validation dataset of BraTS'21 (62 subjects) measured in ($\uparrow$), $\lceil \text{Dice} \rceil$ ($\uparrow$) and $\text{Dice}_{\text{Yen}}$ ($\uparrow$). Each row includes the cumulative additions from the previous rows.}
\label{tab:ablation}
\small
\begin{tabular}{lccc}
 & \multicolumn{3}{c}{BraTS21$_\text{val}$}\\ 
\cmidrule(lr){2-4}
\multicolumn{1}{c}{Method} & AUPRC& $\lceil \text{Dice} \rceil$ &$\text{Dice}_{\text{Yen}}$\\
\midrule
\texttt{ANDi}$_{\text{base}}$&0.221&0.270&0.114\\
~~~~+ Max pooling &0.230&0.283&0.114\\
~~~~+ Geometric mean &0.268&0.319&0.123\\
~~~~+ Pyramidal training&\cellcolor{tabthird}0.674&\cellcolor{tabthird}0.609&\cellcolor{tabthird}0.385\\
~~~~+ MF size 5&\cellcolor{tabsecond}0.853&\cellcolor{tabsecond}0.784&\cellcolor{tabsecond}0.760\\
~~~~+ Eval. with Gauss &\cellcolor{tabfirst}\textbf{0.887}&\cellcolor{tabfirst}\textbf{0.807}&\cellcolor{tabfirst}\textbf{0.784}\\
\bottomrule
\end{tabular}
\end{table}

\section{Discussion \& Conclusion}

In this work, we proposed a novel DDPM-based algorithm for unsupervised anomaly detection in medical imaging data. Our proposed approach functions by directly aggregating deviations of the learned denoising step from the ground truth backwards transition. In contrast to recent related work, our method does not require an additional in-painting or pseudo-healthy image generation step, making it simple and fast to use. In order to achieve state-of-the-art performance we introduced crucial methodological contributions. Most notably, we use pyramidal noise during training of the DDPM, as well as geometric mean for aggregation of the above deviations. 

We further proposed an evaluation strategy based on Yen thresholding of individual anomaly maps ($\text{Dice}_{\text{Yen}}$), which does not require any knowledge of, or data from the target domain. While the established AUPRC evaluation also does not require setting a specific threshold, the advantage of the Yen thresholding is that it can be used to obtain a binary segmentation without knowledge of the target domain. 

When evaluating our approach in a clinically relevant multi-modal brain MR image scenario on three challenging and diverse datasets (BraTS'21, MSSEG, Ljubljana), we clearly outperform two recent diffusion model-based approaches (AnoDDPM \& AutoDDPM), as well as the strong DAE baseline proposed by Kascenas \etal~\cite{dae}. In particular, the detection of MS lesions detection is substantially improved, exceeding the next best method by up to 178\% in terms of AUPRC.

Combined, the developments presented in this paper are a step in the direction of achieving more clinical utility for diffusion model-based UAD. However, it is also clear that while the work is promising, UAD methods have not yet achieved the performance required for clinical utility. In future work, we aim to investigate the interplay of DDPMs with different noise types, which appears to be of great importance for improving the performance of our anomaly detection method. While our method is currently trained on 2D slices, in subsequent work, we aim to better incorporate the inherently three-dimensional nature of brain MRI scans. 

\section{Acknowledgements}
We thank Jakob H. Macke for his helpful comments.
This work was funded by the Deutsche Forschungsgemeinschaft (DFG, German Research Foundation) under Germany's Excellence Strategy – EXC 2064/1, Project number 390727645, with additional support from the DFG Emmy Noether 513851350 (TW) and the German Federal Ministry of Education and Research (BMBF) via the Tübingen AI Center, FKZ: 01IS18039A.
We also utilized the Tübingen Machine Learning Cloud, supported by DFG FKZ INST 37/1057-1 FUGG.
The authors thank the International Max Planck Research School for Intelligent Systems (IMPRS-IS) for their support of Jaivardhan Kapoor, Thomas Wolfers, and Christian F. Baumgartner.

{\small
\bibliographystyle{ieeenat_fullname}
\bibliography{egbib}
}

\clearpage

\appendix

\twocolumn[
\begin{@twocolumnfalse} %
  \begin{center}
    {\Large\bfseries Supplementary Material\par} %
    \vspace{1cm} %
  \end{center}
\end{@twocolumnfalse}
]

\begin{figure}[th]
    \centering
    \includegraphics[width=1\linewidth]{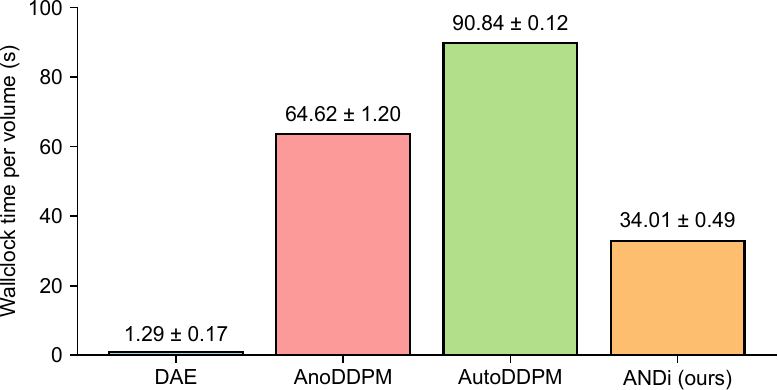}
    \caption{Wallclock time for processing 1 volume ($128 \times 128 \times 155$) for the unsupervised baselines and \texttt{ANDi}. We report the mean $\pm$ standard deviation over 10 runs. \texttt{ANDi} is substantially faster than the other DDPM-based methods, and is only outperformed by the simpler DAE. }
    \label{fig:wallclock}
\end{figure}

\section{Training and Evaluation Hyperparameters}
\label{appendix:a}
All DDPM models were parameterized by the same U-Net architecture.
\texttt{ANDi} was trained with pyramidal noise, AutoDDPM with Gaussian noise, and AnoDDPM with simplex noise. All hyperparameters for the simplex noise are equal to the original publication with a starting frequency of $\nu = 2^{-6}$
, octave of $N = 6$ and a decay of $\gamma = 0.8$. When generating the simplex
noise, the seed was changed before every calculation,
and a slice $s$ from the 3-dimensional noise function was extracted. We trained \texttt{ANDi} with a total number of diffusion timesteps $T=1000$, a linear noise schedule between $\beta_1 = 10^{-4}, \beta_T = 0.02$ and a batch size of $128$ over $232$ epochs. Starting with a learning rate of $2 \cdot 10^{-5}$, we linearly increased it to $1 \cdot 10^{-4}$ over 5\% of the total training steps and then decayed it back to $2 \cdot 10^{-5}$ with a cosine schedule.

All diffusion model-based methods require setting an upper ($T_u$) and a lower bound ($T_l$) for the denoising operations during evaluation. For \texttt{ANDi} we empirically determined $T_u=200$ to $T_l=75$ to work well, though we show that \texttt{ANDi} is not sensitive to the exact choice of time range in \cref{appendix:c}. For AutoDDPM and AnoDDPM we followed the original publications when setting the time range. AutoDDPM and AnoDDPM require $T_l=0$ as they need denoising back to the image space fully. The initial mask generation step of AutoDDPM uses $T_u=200$ and AnoDDPM uses $T_u=250$. AutoDDPM requires an additional thresholding hyperparameter during evaluation to obtain the binary mask for in-painting. In contrast to the original publication, this threshold was set to the 99th percentile of the residuals obtained on the validation dataset of BraTS'21. This choice improved the results for AutoDDPM. The subsequent inpainting process was run for the last $50$ denoising steps with $5$ resamplings per step.

\begin{figure}[t]
    \centering
    \includegraphics[width=1.0\linewidth]{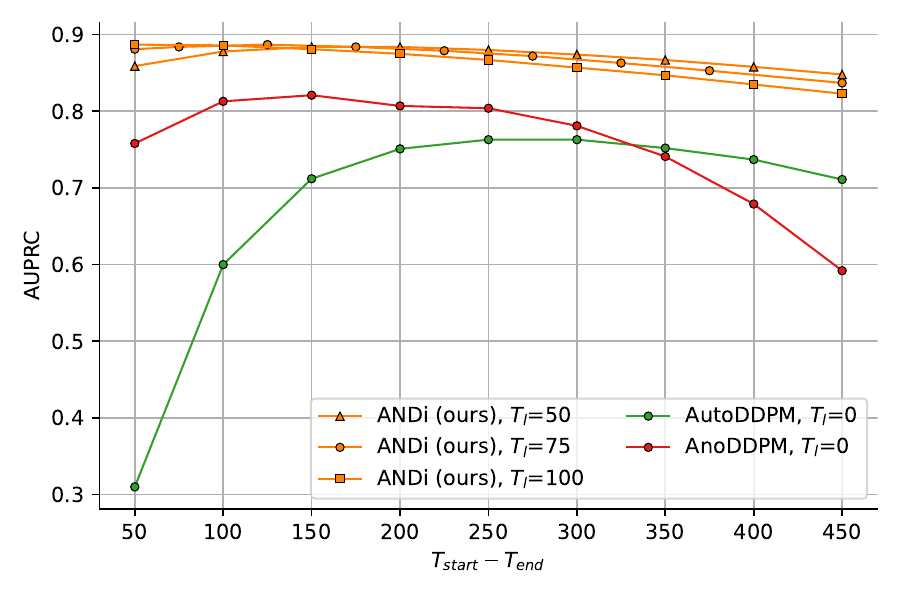}
    \caption{Influence of the chosen time step interval ($T_u - T_l$) on the performance measured in AUPRC for \texttt{ANDi} and the Diffusion Model baselines. \texttt{ANDi} is the most robust method regarding this hyperparameter choice and reaches optimal performance with the smallest amount of denoising steps.}
    \label{fig:time_step}
\end{figure}

\section{Wallclock Analysis of Methods}
\label{appendix:b}
In order to better understand the computational time required for anomaly detection using the different methods, we performed a wallclock analysis for one volume. 

\cref{fig:wallclock} shows the wallclock time for processing the anomaly maps for each of the unsupervised baselines, as well as \texttt{ANDi}. This analysis includes postprocessing steps over the anomaly maps, such as maxpool over modalities, median filtering, and thresholding. These steps are common across all methods and, therefore, contribute a constant offset $1.08 \pm 0.02 s$ to the wallclock time. We used an NVIDIA-V100 GPU on a 16-core machine for all evaluations.

\texttt{ANDi} is the fastest among the DDPM-based methods and can process a volume almost twice as fast as the next best method in that category, AnoDDPM. \texttt{ANDi} is only outperformed by the DAE baseline, which predicts the anomaly map in a single pass of the UNet. 

The speedups of \texttt{ANDi} with respect to the other DDPM-based methods can largely be explained due to the number of denoising steps considered in the evaluation. \texttt{ANDi} requires fewer timesteps than AnoDDPM and AutoDDPM to achieve maximum performance (see also \cref{fig:time_step}). Further, the gains with respect to AutoDDPM can be explained by the fact that AutoDDPM requires two stages for anomaly detection, an initial anomaly mask creation and a subsequent inpainting stage, while \texttt{ANDi} requires only a single stage.

In our evaluations of AnoDDPM, in order to reduce the computational complexity, we reused the same simplex noise sample for each slice of a test volume, but augmented it with random rotations and flipping. Computing independent simplex noise for each slice of a volume would take significantly longer due to the need to sample 3D simplex noise to reduce corner artifacts. The noise samples are still independent for each step of the diffusion process. 
If implemented with slice-wise independent simplex noise, AnoDDPM would be substantially slower, and would likely fall behind AutoDDPM in terms of execution time. 

We note that \texttt{ANDi} has the potential to be parallelized over denoising timesteps, given adequate GPU memory and FLOPs. While we didn't explore this in the present paper, parallelization over denoising steps would lead to further substantial speedups. In contrast, AnoDDPM and AutoDDPM are sequential by nature and hence cannot be parallelized in this way.

\section{Sensitivity Analysis on Number of Denoising Timesteps}
\label{appendix:c}
As mentioned earlier, all DDPM-based algorithms required setting an upper ($T_u$) and a lower bound ($T_l$) for the denoising operations during evaluation. The resulting time range $T_u - T_l$ is an important hyperparameter for all methods. 

AnoDDPM and AutoDDPM require setting $T_l=0$ because they need to denoise back to the input space. \texttt{ANDi}, on the other hand, is not required to reach the end of the diffusion chain because no pseudo-healthy image is not calculated. Therefore, we can choose any $T_l$ as an endpoint for the denoising steps.

In order to better understand the DDPM-based algorithms' sensitivity to the time range $T_u - T_l$, we performed a parameter exploration on the validation dataset of BraTS'21. The results are shown in \cref{fig:time_step}. We observe that \texttt{ANDi} is significantly more robust to the choice of $T_u$ and $T_l$. Moreover, we note that all \texttt{ANDi} configurations within our exploration range outperform the best AnoDDPM and AutoDDPM configurations. It can further be observed that ANDi obtains optimal anomaly detection with a smaller number of total denoising steps compared to the other two methods. This partially explains the speedups we measured in \cref{appendix:b}.

\section{Additional Randomly Sampled Qualitative Results}

\begin{figure*}[h]
    \centering
    \includegraphics[width=\linewidth]{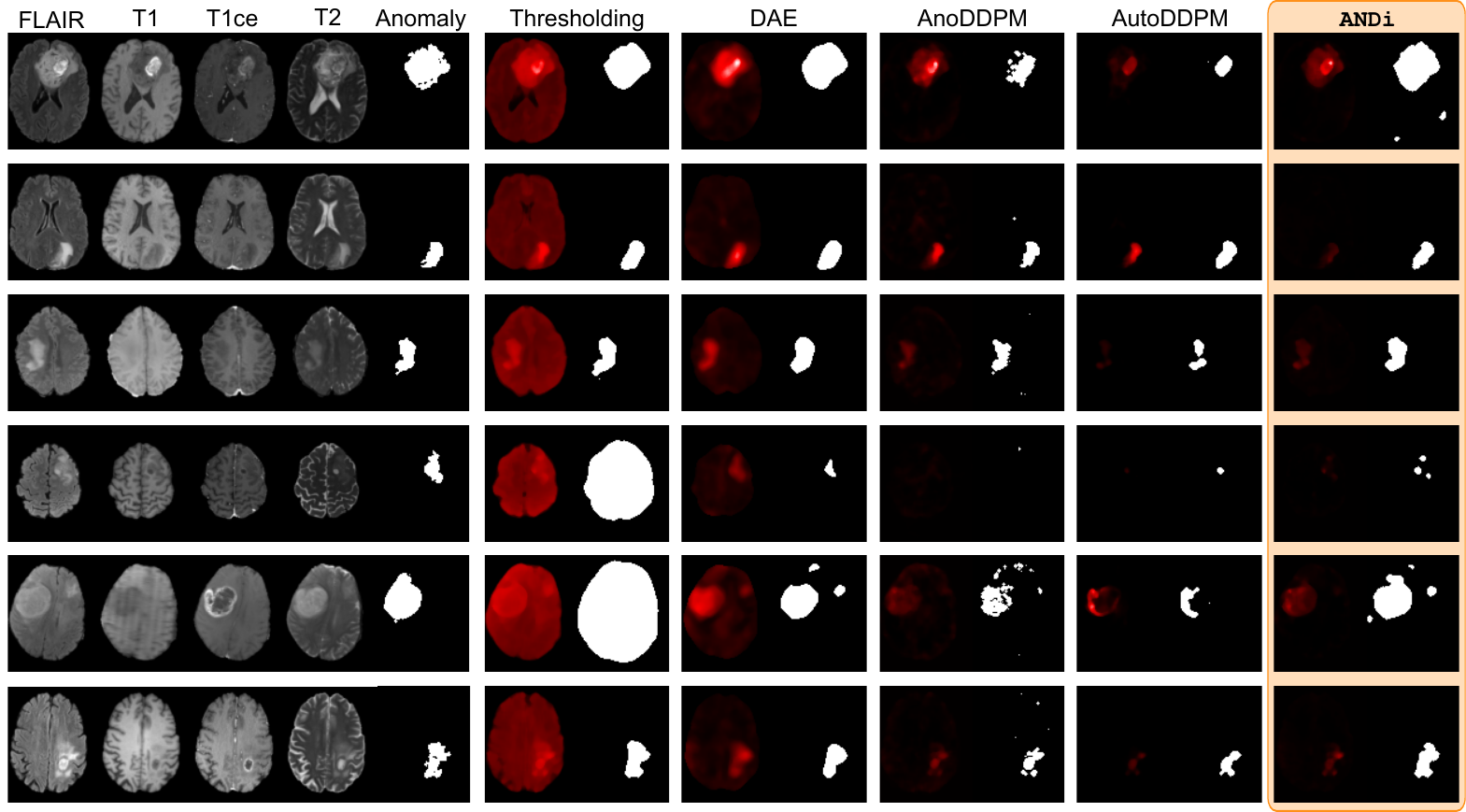}
    \caption{Randomly selected qualitative results for anomaly maps and Yen thresholding on the BraTS'21 test dataset. The anomaly maps are post-processed using median filtering (kernel size 5).}
    \label{qual:brats}
\end{figure*}

\begin{figure*}[th]
    \centering
    \includegraphics[width=\linewidth]{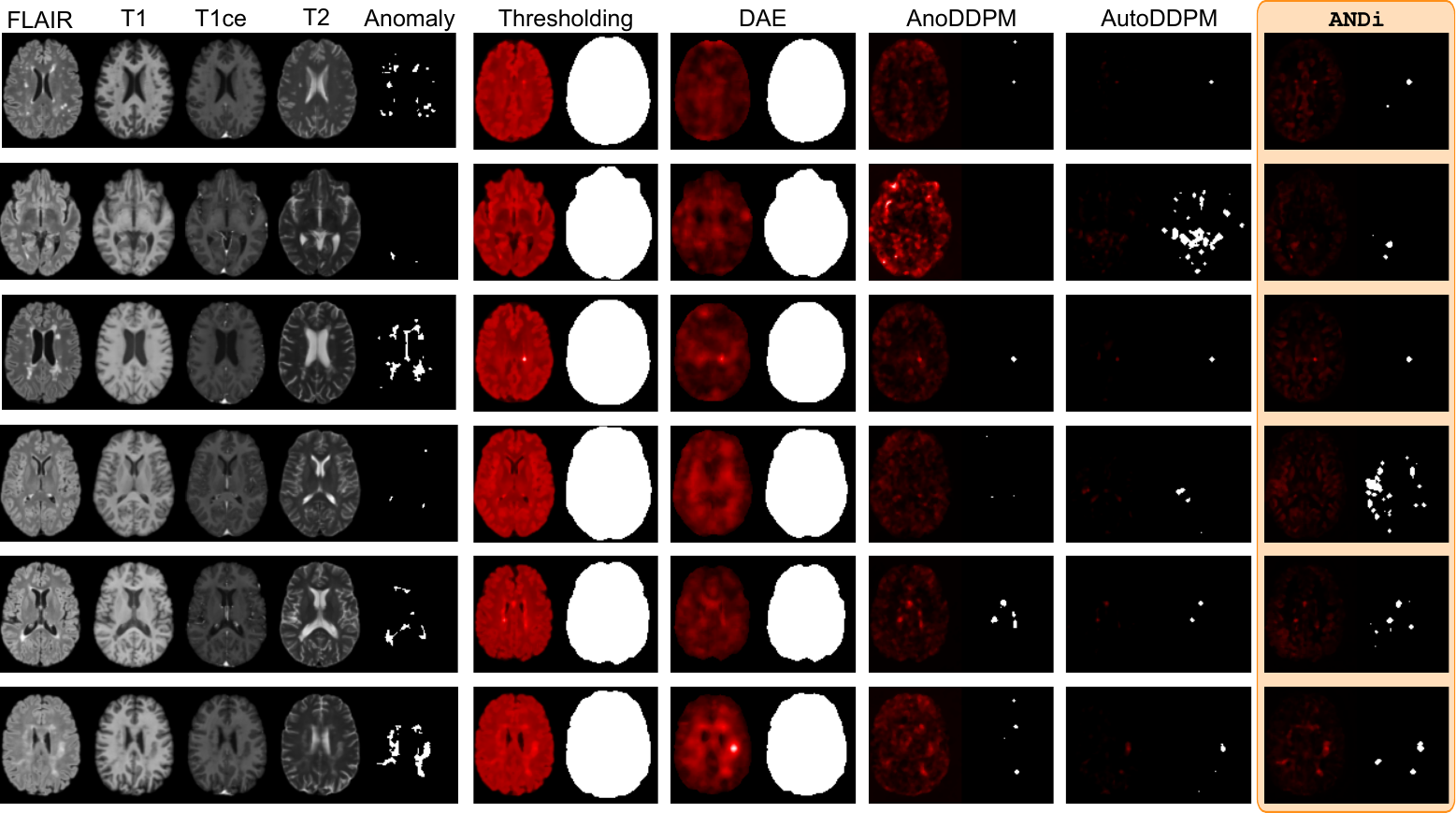}
    \caption{Randomly selected qualitative results for anomaly maps and Yen thresholding on the Shifts-MSSEG test dataset. The anomaly maps are post-processed using median filtering (kernel size 3).}
    \label{qual:shifts1}
\end{figure*}

\begin{figure*}[th]
    \centering
    \includegraphics[width=\linewidth]{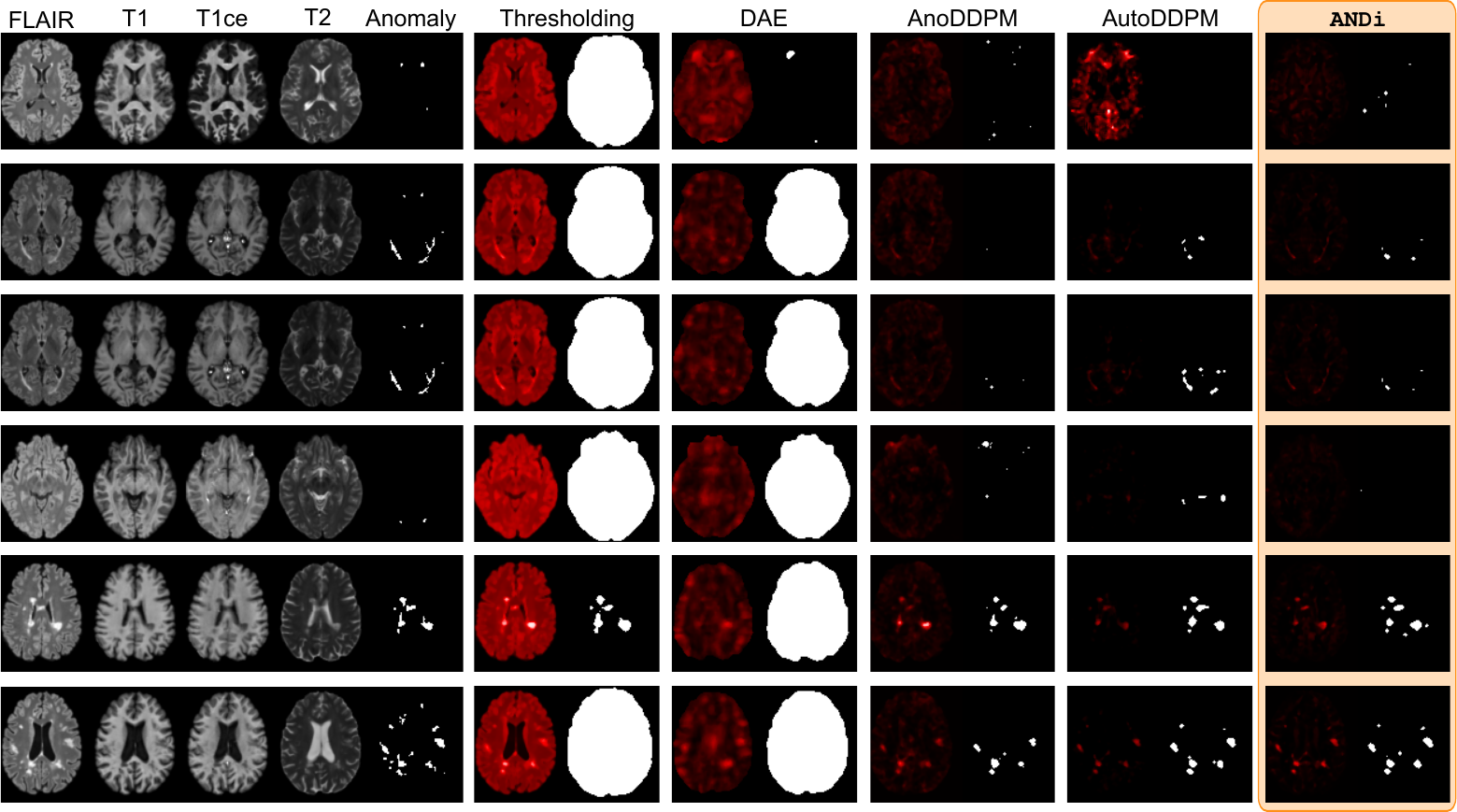}
    \caption{Randomly selected qualitative results for anomaly maps and Yen thresholding on the Shifts-Ljubljana test dataset. The anomaly maps are post-processed using median filtering (kernel size 3).}
    \label{qual:shifts2}
\end{figure*}

In the following we present additional qualitative anomaly detection results for all methods on the BraTS'21 (\cref{qual:brats}), MSSEG (\cref{qual:shifts1}), and Ljubljana datasets (\cref{qual:shifts2}). As in the main article, we show the anomaly maps and the binarized segmentations obtained using Yen thresholding. The shown images were randomly sampled from the test datasets. 

The results shown use median filtering with a filter size of $5\times 5\times 5$ for the BraTS21 dataset and $3\times 3\times 3$ for the two Shifts datasets. As mentioned in the main text, a larger kernel size hurts performance for small anomalies, such as MS lesions in the Shifts dataset.

For MS lesions, the DAE and the Thresholding approaches often detect the complete slice as anomalous when using Yen thresholding. This happens when the volume is segmented into foreground and background due to the anomaly maps not being informative enough to find an appropriate threshold. %

\end{document}